\def\year{2022}\relax
\documentclass[letterpaper]{article} 
\usepackage{aaai22}  
\usepackage{times}  
\usepackage{helvet}  
\usepackage{courier}  
\usepackage[hyphens]{url}  
\usepackage{graphicx} 
\usepackage{tablefootnote}
\usepackage{natbib}  
\usepackage{caption} 
\DeclareCaptionStyle{ruled}{labelfont=normalfont,labelsep=colon,strut=off} 
\usepackage{algorithm}
\usepackage{algorithmic}
\usepackage{booktabs}
\usepackage[dvipsnames]{xcolor}
\usepackage[caption=false]{subfig}
\usepackage{newfloat}
\usepackage{listings}
\floatstyle{ruled}
\newfloat{listing}{tb}{lst}{}
\floatname{listing}{Listing}
\title{Curator: Creating Large-Scale Curated Labelled Datasets using Self-Supervised Learning}
\author{
    Tarun Narayanan\equalcontrib \textsuperscript{\rm 1}, Ajay Krishnan\equalcontrib \textsuperscript{\rm 1}, Anirudh Koul\textsuperscript{\rm 1,2,4}, Siddha Ganju\textsuperscript{\rm 1,3,4}
    }
\affiliations{
    \textsuperscript{\rm 1}SpaceML
    \textsuperscript{\rm 2}Pinterest 
    \textsuperscript{\rm 3}NVIDIA
    \textsuperscript{\rm 4}Frontier Development Lab \\
   tarunn2799@gmail.com, krishajay.g@gmail.com
}

\usepackage{bibentry}

\begin{document}

\maketitle

\begin{abstract}
Applying Machine learning to domains like Earth Sciences is impeded by the lack of labeled data, despite a large corpus of raw data available in such domains. For instance, training a wildfire classifier on satellite imagery requires curating a massive and diverse dataset, which is an expensive and time-consuming process that can span from weeks to months. Searching for relevant examples in over 40 petabytes of unlabelled data requires researchers to manually hunt for such images, much like finding a needle in a haystack. We present a no-code end-to-end pipeline, Curator \footnote{We release all instructions, trained models and code for Curator : https://www.github.com/spaceml-org}, which dramatically minimizes the time taken to curate an exhaustive labeled dataset. Curator is able to search massive amounts of unlabelled data by combining self-supervision, scalable nearest neighbor search, and active learning to learn and differentiate image representations. The pipeline can also be readily applied to solve problems across different domains. Overall, the pipeline makes it practical for researchers to go from just one reference image to a comprehensive dataset in a diminutive span of time.
\end{abstract}

\section{Introduction} One of the initial steps for a scientific study related to climate change and natural disasters, including wildfires, oil spills, hurricanes, dust storms, etc., involves scientists gathering a large number of relevant examples from satellite imagery. Locating an exhaustive set of examples requires painstakingly inspecting 197 million square miles of satellite imagery each day across more than 20 years. While such an effort can produce a valuable trove of data, the act of manually searching is laborious, expensive, and often impractical - grounding many scientific studies before they could ever take off. 

While one of the approaches to solving this is building an image similarity search, several challenges arise when applying similarity search to raw satellite imagery:
\begin{itemize}
    \item The data is unlabelled, preventing attempts to train conventional supervised models which could have generated meaningful representations.
    \item Pretrained ImageNet \cite{5206848} models fail to transfer representations and generalize to this data - especially for larger areas that are usually without sharp edges including clouds as well as multi-spectral data.
    \item Climate phenomena can have vastly different physical sizes - from few miles for wildfires to 300+ miles for hurricanes. 
    \item Vast data imbalances inherently present in the data.
    \item The engineering challenges that come with the sheer scale of our data.
\end{itemize}

We propose Curator, a modular toolkit that aims to take a user from one reference image to an exhaustive set of relevant examples for any large unlabelled image data source. It solves the core issue of data inaccessibility by discovering relevant samples from sizeable collections while minimizing human labeling effort. This pipeline combines several individually tested, high-performance components built for specific tasks - from downloading data, training self-supervised models, large-scale similarity search, active learning, and crowd-sourced labeling. This open source project, built by citizen scientists \cite{DBLP:journals/corr/abs-2012-10610}, aims to enable a researcher to accomplish all of this without writing a single line of code or possessing any prerequisite AI knowledge. This ease of usage further reduces barriers to entry and hopefully catalyzes research involving climate science.

\begin{figure*}[t]
\centering
\includegraphics[width=1.0\textwidth]{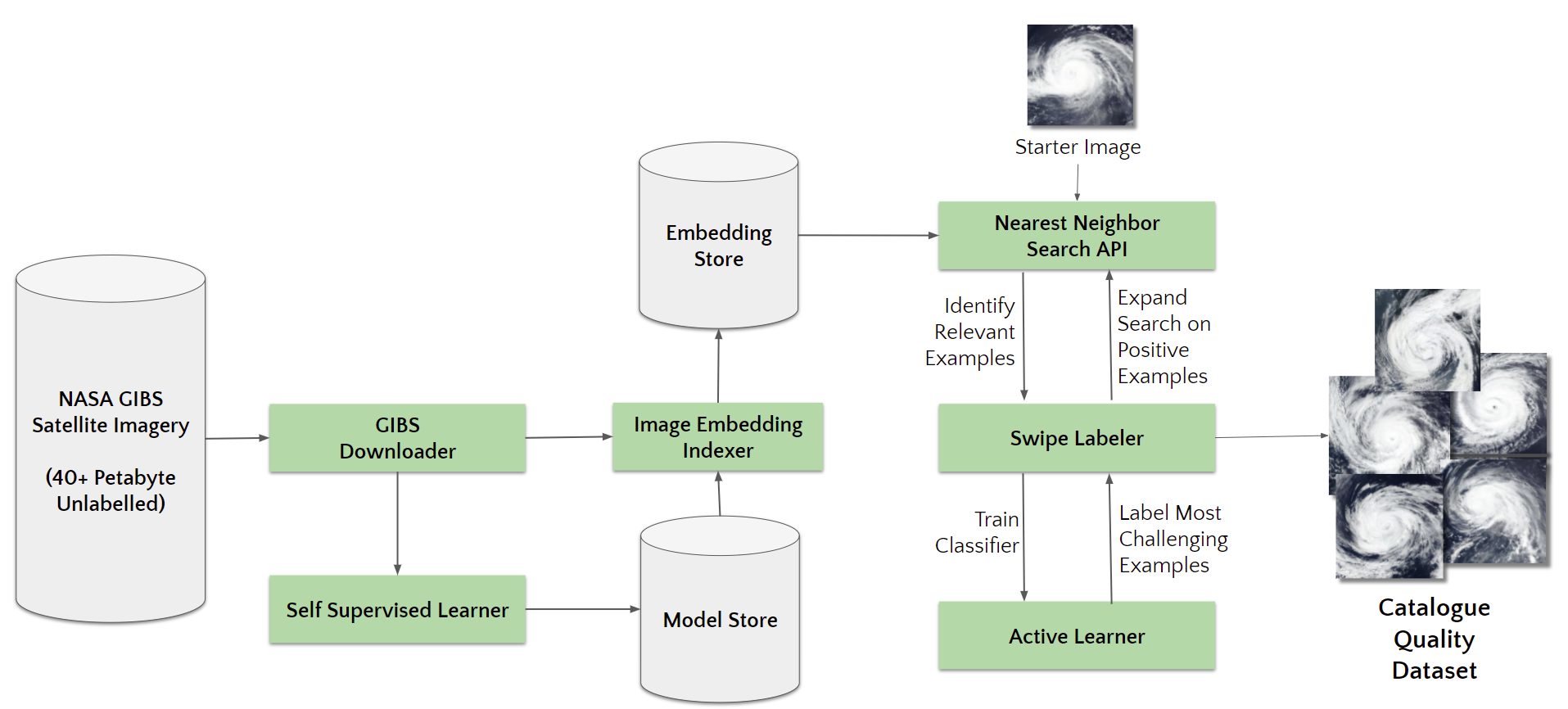} 
\caption{The Curator pipeline}
\label{fig2}
\end{figure*}

\section{Previous Methods}
We demonstrate a specific use case of our pipeline that aims to solve a previously unsolved problem - building a curated dataset for any natural phenomenon by intelligently indexing over 897 satellite imagery sources via the Global Imagery Browse Services (GIBS) portal. To the best of our knowledge, the widely accepted solution in practice which also acts as the baseline is a manual approach involving visual inspection of data from the GIBS portal for multiple layers over a region for a period of time, where each layer provides information overlays based on science disciplines, hazards, and disaster categories, downloading the requisite data, and then manually annotating it. Research in this field includes manually labeling, semi-supervised learning like in \cite{JTD31214}, or using text mining and NLP techniques to extract images and their labels from multiple large data stores. Our method involves annotating a negligible number of images in comparison, and then relies on active learning to generate weak labels for the rest of the dataset.

\section{Pipeline}
Our key goal is to let a scientist use a single query image (say of a climate event) to ultimately identify every potential example of the same category in a large image collection (like satellite imagery).  A scalable way to do this is by evaluating each image with a classifier tuned to the user’s needs. Training such a supervised classifier requires enough positive and negative examples for training. Getting to this training set can be achieved in four steps - \textbf{(1)} training a self-supervised model on unlabeled data, in order to learn semantically relevant representations. \textbf{(2)} generating embeddings for the entire dataset \textbf{(3)} for one or more starter examples,  building a seed set of similar images, i.e images with embeddings corresponding to the nearest neighbours to the query image \textbf{(4)} using several iterations of human-in-the-loop active learning to find examples that maximize classifier performance while minimizing human labeling time. 

\begin{table*}[]
    \centering
    \begin{tabular}{l p{1cm}*{8}{p{1.5cm}}}
        \toprule
        Active Learning Strategy & F1 Score (Val) & Total labelling effort by the user & Positive Images Retrieved & False Positive Images Retrieved \\
        \midrule
         Random Sampling (with Imagenet Pretraining) & 0.45 & 7.6\% & 65\% & 37\% \\
         Uncertainty Sampling (with SSL Pretraining) & \textbf{0.74} & 7.8\% & \textbf{88\%} & 12\% \\
        \bottomrule
    \end{tabular}
    \caption{Number of positive images along with the percentage of data predicted as False Positives, that were retrieved across different active learning strategies.}
    \label{tab:experiment1}
\end{table*}

The modules of Curator can be combined to achieve this functionality (summarized in Fig 1). Key themes in their development include that each tool need to be 1) executable through a single command 2) highly modular so it can be used for an individual task or combined for a range of tasks, including beyond climate science 3) built for high performance with the available hardware (single, multiple GPU or multi node) while being cost effective at scale. With the aim to get researchers started in minutes, the tools can be run on a local machine through a simple command line interface. For higher scale, the pipeline provides a cloud specific template using Google Cloud (which can be replicated but needs deeper familiarity with the cloud). We also include a set of data preprocessing functions that were designed to solve some inherent deficiencies present in the satellite imagery data (for more information see Appendix A).

\subsection{GIBS Downloader}

GIBS Downloader \cite{lisboa2020democratizing} is a command-line tool that simplifies access to satellite imagery from NASA Global Imagery Browse Services (GIBS), thereby tackling all the esoteric challenges behind acquiring and processing decades of satellite imagery data. It provides access to over 897 products, along with the ability to search their remote sensing product descriptions by keywords. It offers various functionalities to easily convert datasets to a format that can be directly used for AI training, including TensorFlow's TFRecords for accelerating the speed of data ingestion in training pipelines. The required arguments include the date range and the lat/long coordinates of the rectangular region. Operating on a canvas of up to 262144 x 131072 pixels for a full view of the globe (which cannot be opened by most image viewers), it uses several performance optimizations like multithreading to parallelize extraction of smaller tiles suited for a researcher’s needs. 

\subsection{Self Supervised Learner}

Self Supervised Learner is a command-line tool that takes a directory of unlabeled images and trains a self-supervised model. Self-Supervised Learning (SSL) is a relatively new method of unsupervised representation learning wherein we generate temporary labels intrinsically from the images by exposing a relationship between different parts of the image or with multiple views of the image. Currently, the SimCLR \cite{chen2020simple}, and the SimSiam \cite{chen2021exploring} architectures are supported. Built for performance, the Self-Supervised Learner utilizes NVIDIA DALI package to parallelize CPU operations like image decoding and augmentations on the GPU, resulting in up to 8x speedup in training time. The tool can scale training from single GPU to multi GPU, consistently with 90\% GPU resource utilization off-the-shelf. It also provides a high level of customizability in defining custom model architectures, augmentations, along with planned support for multi-band data and seasonal contrast modeling \cite{manas2021seasonal}. 

\subsection{Scalable Image Search}

Curator provides a command-line tool for local machines as well as a Google Cloud template to perform scalable interactive image search.  First, the Image Embedding Indexer takes a model and generates embeddings rapidly (through GPU acceleration using NVIDIA DALI). Then, these embeddings are indexed for fast approximate nearest neighbor search using FAISS \cite{DBLP:journals/corr/ChengHL17}. Lastly, a low latency API provides image query capabilities along with filtering options. Additionally, the pipeline provides an interactive UI to visualize search results. The search index is partitioned by date, resolution, and product to make the system scalable and parallelizable. For an image collection with up to 5 million images, most modern laptops can retrieve results in under a second, satisfying the requirements of most researchers and enabling them to get started quickly. For larger collections, the cloud template contains several performance tweaks to parallelize and run a scalable yet cost-efficient multi-node system, such as utilizing Google Google Cloud Functions, reading the index files as a byte stream, configuring the same regions for bucket and VM regions, and more. 

\subsection{Swipe Labeler}
Swipe Labeler is a browser-based annotation tool meant to quickly and efficiently label image collections with binary labels. It is intended to make the usually tedious process of labeling data more engaging by swiping right or left (or pressing right/left arrow keys) to move the images into folders categorized as relevant and non-relevant. Accessible on both mobile and desktop, the tool can be activated by a single command. The tool offers multi-user collaborative labeling by seamlessly generating a public shareable link without the user requiring any networking knowledge.

\subsection{Active Labeler}

Active Labeler (AL) is a tool that incorporates human-in-the-loop active learning to minimize labeling while maximizing classifier performance. Given a seed set of labeled images, it trains a classifier (transfer learning on the SSL model, or any image classification setup), evaluates all unlabeled images and picks a small subset for human labeling, which are added to the labeled image set. It repeats this process iteratively till the classifier shows robust performance metrics. A variety of strategies can be employed to identify the data points that would contribute most to the accuracy of the
model, in other words, they calculate which data points are most ’influential’. The tool supports a range of strategies fundamentally based on Uncertainty Sampling such as Least Confidence Sampling, Margin Based Sampling, and Entropy Based Sampling. With a sampling strategy that is based only on uncertainty, there is a possibility that the samples
selected for training are very similar to each other. In such a scenario, intuitively, the model would
only learn about a certain type of image in each iteration, rendering the process inefficient. The
inclusion of diversifying sampling strategies may help fully utilize each iteration, ensuring that the
model learns from a set of diverse samples as opposed to a homogeneous one. The strategies that have been implemented thus far are Iterative proximity-based sampling, Gaussian Sampling and Clustering-based sampling. Beyond basic active learning, AL also interfaces with Scalable Image Search. It helps build a labeled seed set by taking a single starter image, retrieving similar images, and labeling them with Swipe Labeler. The seed images should contain distinguishable features that you want to distinctively see in the retrieved similar images. At scale, several performance tweaks have been incorporated, including - \textbf{(1)} using embeddings instead of images to significantly reduce computation \textbf{(2)} training a classification head using features from a pretrained SSL backbone \textbf{(3)} reducing the output dimension of the SSL backbone, to improve downstream training time and space efficiency. 
Additionally, we utilize a subsample of approximately equidistant embedding vectors (Core-Set) instead of the entire embedding space in order to exponentially reduce the time taken to perform a forward pass operation (for more, refer Appendix A). The datapoints selected in the subsample are then used to find the nearest neighbors in the entire embedding space. With these improvements, leveraging multi-million to billion scale image datasets becomes practical from a cost and latency standpoint.

\begin{figure*}[t]
\centering
\includegraphics[width=0.9\textwidth]{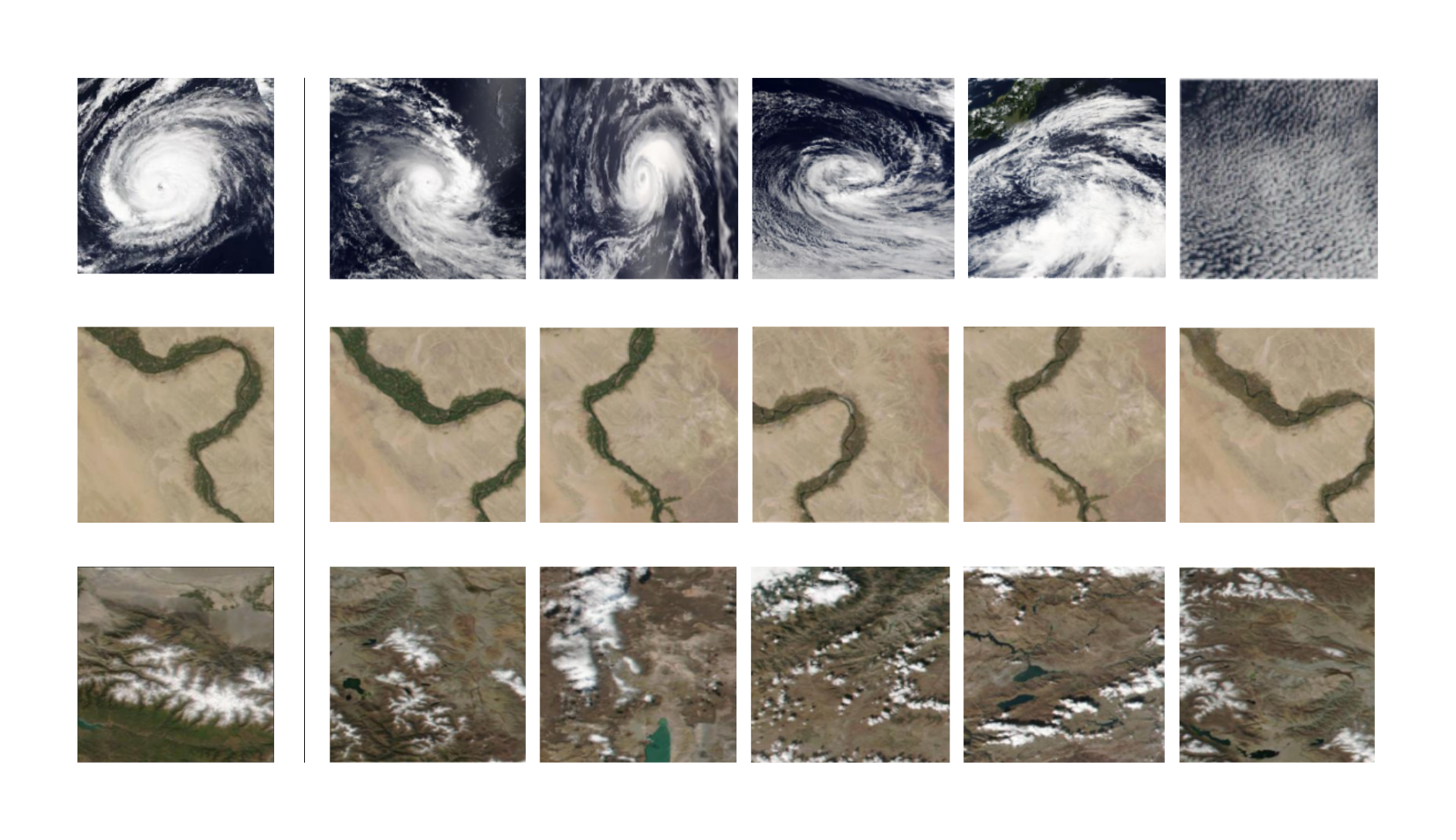} 
\caption{Image Retrieval results on VIIRS data. (Left) Query Image (Right) Retrieved images from the curated set.}
\label{fig3}
\end{figure*}

\section{Results}

To evaluate the effectiveness of the pipeline on a labeled benchmark dataset containing satellite imagery, we experimented with RESISC45 \cite{DBLP:journals/corr/ChengHL17} (Remote Sensing Image Scene Classification), which contains 31,500 images, covering 45 classes with 700 images in each class with high intra-class diversity and inter-class similarity, making it relatively challenging. Given a single reference image, we aim to evaluate the number of images of the same class that can be identified, along with the amount of human labeling required. For a starter image, a seed set is constructed and then assigned positive/negative class labels. This seed set consists of 64 nearest neighbors to the starter image and 32 randomly sampled images to provide a diverse negative class. This seed set is used by the Active Labeler, which iteratively trains a classifier, classifies the entire dataset, and picks a subset of 64 most informative images to be assigned a label, which is then used in the subsequent iteration for training. The system runs till 5\% of the dataset has been labeled. The resulting classifier is then used to identify potential positive classes in the dataset and presented to the user for verification to build a curated set. We repeat the experiment for all 45 classes, with 10 randomly chosen starter images per class. Results, shown in Table~\ref{tab:experiment1}, showcase that, on average, 88\% of the images belonging to the same class as the starter image was retrieved with 7.8\% manual labeling effort. This result is in contrast to the baseline of manually evaluating every single image in the dataset.

To further battle test our pipeline in real-time data scenarios, we setup Curator to curate images from an unlabeled satellite imagery dataset. We tiled and retrieved 30 days' worth of data from the VIIRS product using the GIBS Downloader tool, and we pretrained SimCLR on this data using relevant augmentation strategies for 1000 epochs on a single GPU. This model is the backbone for Active Labeler. We picked starter images from our validation set and passed them to Curator to retrieve similar images. Examples of starter images and images from their curated set are illustrated in Figure 2. 

We believe another important outcome of using our pipeline is the underlying time and monetary benefit that comes from rapid iteration. For example, let's evaluate the task of finding images of islands from NASA Worldview. During a recent demonstration of Curator on the NASA GIBS/Worldview imagery pipeline, a machine was trained to search for islands through five million tiles of Earth imagery starting with a single seed image of an island. Approximately 1,000 islands were identified in just 52 minutes with just one human in the loop. If done manually, this effort would take an estimated 7,000 hours (assuming five seconds to evaluate and label each image tile) and potentially cost as much as \$105,000 (assuming \$15 per hour per annotator) \cite{nasablog}.

\section{Conclusion} We present a novel pipeline that provides an automated approach to curating relevant datasets starting from a single image with significantly less human effort involved. Built for scale and cost effectiveness, the pipeline leverages techniques like self-supervised learning, human-in-the-loop active learning, geometric data sampling, and nearest neighbor search. Reducing the time of manual data curation from several months to hours or even minutes opens new avenues of scientific exploration previously considered impractical. By releasing a readily usable open-source toolbox, we hope to accelerate research in domains like climate science, where access to structured data and has been a major challenge.

\bibliography{references.bib}

\begin{thebibliography}{10}
\providecommand{\natexlab}[1]{#1}

\bibitem[{Blumenfeld(2021)}]{nasablog}
Blumenfeld, J. 2021.
\newblock SpaceML: Rise of the Machine (Learning).

\bibitem[{Chen et~al.(2021)Chen, Cao, Koul, Ganju, Praveen, and
  Kasam}]{chen2021reducing}
Chen, S.; Cao, E.; Koul, A.; Ganju, S.; Praveen, S.; and Kasam, M.~A. 2021.
\newblock Reducing Effects of Swath Gaps on Unsupervised Machine Learning
  Models for NASA MODIS Instruments.
\newblock \emph{arXiv preprint arXiv:2106.07113}.

\bibitem[{Chen et~al.(2020)Chen, Kornblith, Norouzi, and
  Hinton}]{chen2020simple}
Chen, T.; Kornblith, S.; Norouzi, M.; and Hinton, G. 2020.
\newblock A simple framework for contrastive learning of visual
  representations.
\newblock In \emph{International conference on machine learning}, 1597--1607.
  PMLR.

\bibitem[{Chen and He(2021)}]{chen2021exploring}
Chen, X.; and He, K. 2021.
\newblock Exploring simple siamese representation learning.
\newblock In \emph{Proceedings of the IEEE/CVF Conference on Computer Vision
  and Pattern Recognition}, 15750--15758.

\bibitem[{Cheng, Han, and Lu(2017)}]{DBLP:journals/corr/ChengHL17}
Cheng, G.; Han, J.; and Lu, X. 2017.
\newblock Remote Sensing Image Scene Classification: Benchmark and State of the
  Art.
\newblock \emph{CoRR}, abs/1703.00121.

\bibitem[{Deng et~al.(2009)Deng, Dong, Socher, Li, Li, and Fei-Fei}]{5206848}
Deng, J.; Dong, W.; Socher, R.; Li, L.-J.; Li, K.; and Fei-Fei, L. 2009.
\newblock ImageNet: A large-scale hierarchical image database.
\newblock In \emph{2009 IEEE Conference on Computer Vision and Pattern
  Recognition}, 248--255.

\bibitem[{Kim et~al.(2019)Kim, Yi, Hager, and Lin}]{JTD31214}
Kim, T.~K.; Yi, P.~H.; Hager, G.~D.; and Lin, C.~T. 2019.
\newblock Refining dataset curation methods for deep learning-based automated
  tuberculosis screening.
\newblock \emph{Journal of Thoracic Disease}, 12(9).

\bibitem[{Koul et~al.(2020)Koul, Ganju, Kasam, and
  Parr}]{DBLP:journals/corr/abs-2012-10610}
Koul, A.; Ganju, S.; Kasam, M.; and Parr, J. 2020.
\newblock Space {ML:} Distributed Open-source Research with Citizen Scientists
  for the Advancement of Space Technology for {NASA}.
\newblock \emph{CoRR}, abs/2012.10610.

\bibitem[{Lisboa et~al.(2021)Lisboa, Verma, Koul, Kasam, and
  Ganju}]{lisboa2020democratizing}
Lisboa, F.; Verma, S.; Koul, A.; Kasam, M.~A.; and Ganju, S. 2021.
\newblock Democratizing Earth Science Research with Accessible Data
  High-Performance Training Pipelines.
\newblock \emph{Committee on Space Research Cloud Computing Workshop}.

\bibitem[{Ma{\~n}as et~al.(2021)Ma{\~n}as, Lacoste, Giro-i Nieto, Vazquez, and
  Rodriguez}]{manas2021seasonal}
Ma{\~n}as, O.; Lacoste, A.; Giro-i Nieto, X.; Vazquez, D.; and Rodriguez, P.
  2021.
\newblock Seasonal Contrast: Unsupervised Pre-Training from Uncurated Remote
  Sensing Data.
\newblock \emph{arXiv preprint arXiv:2103.16607}.

\end{thebibliography}

\appendix
\section{Appendix A: Adapting to Different Tasks}
\addcontentsline{toc}{section}{Appendix A: Adapting to different downstream tasks}
\renewcommand{\thesubsection}{\Alph{subsection}}

The pipeline is generalizable on any unlabelled source of data in a domain agnostic manner. Performing this task simply requires us to define a custom Data Source, Data Downloader, and an optional Data Preprocessor that is specific to the problem we’re solving.

\subsection{Data Source}
The Data Source is a user-provided pool of unlabelled data. Most domains have a lot of data being collected that currently do not translate to value in our context due to their lack of organization, and Curator is designed to leverage these data sources without the hard requirement for annotation. In our demonstration we pick the NASA Worldview platform as our Data Source, and we demonstrate how our pipeline can be used to generate curated datasets from the satellite imagery data available on this platform. 

\begin{figure}
  \centering
  \includegraphics[scale = 0.22]{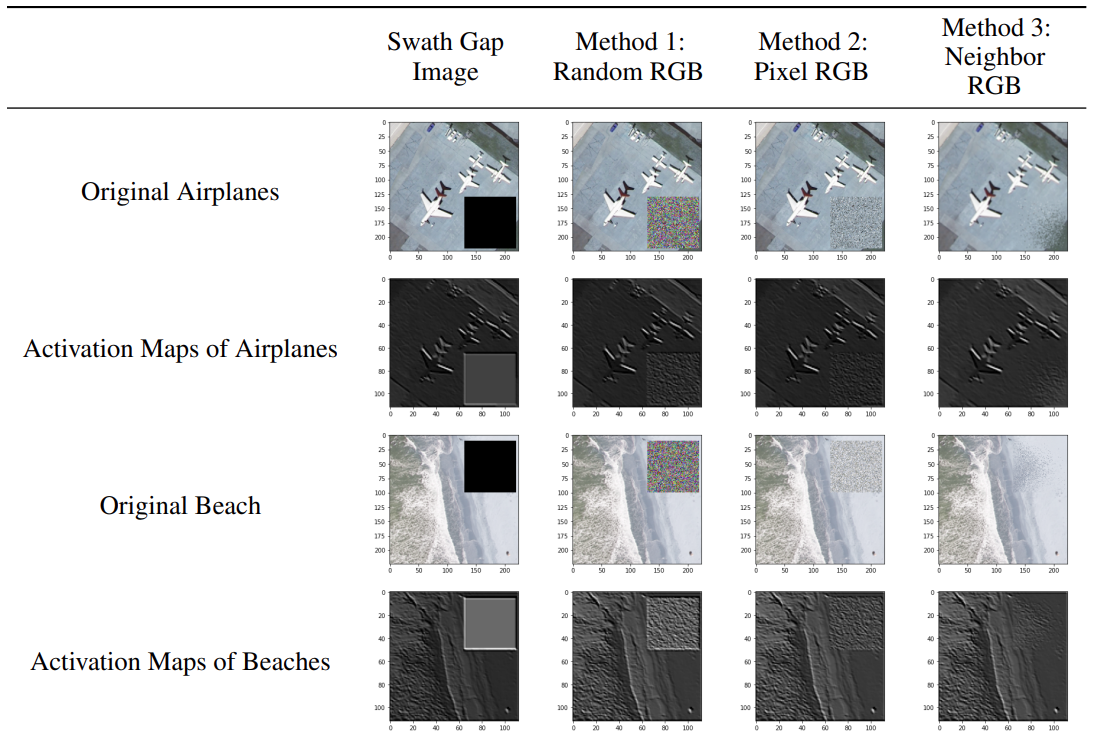}
  \caption{Trained convolutional autoencoder outputs for Swath Filler. Query image (leftmost column) and its corresponding most-similar four images. Filling strategy changes row wise: no fill, Random RGB, Pixel RGB, Neighbor RGB. Random RGB fill strategy results show that the autoencoder focuses on swath gap positions. Neighbor RGB fill strategy results show that the autoencoder ignores the swath gap and concentrates on the ROI.}
\end{figure}

\begin{figure*}
    \centering
    \subfloat[Left: Gulf of Mexico without Clouds generated based on previously available data. Right: Generated cloud masks over the Alps region. ]{{\includegraphics[clip, width=11cm]{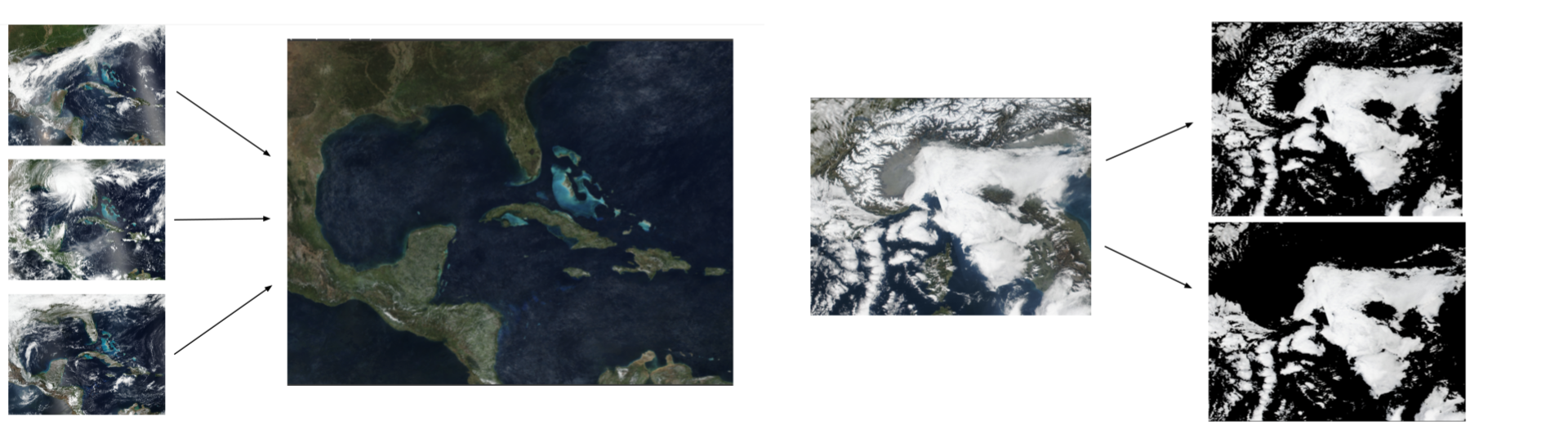} }}
    \qquad
    \subfloat[Image retrieval across multiple resolutions for our Tile-based multi-resolution search against Image-based multi-resolution search]{{\includegraphics[clip, width=11cm]{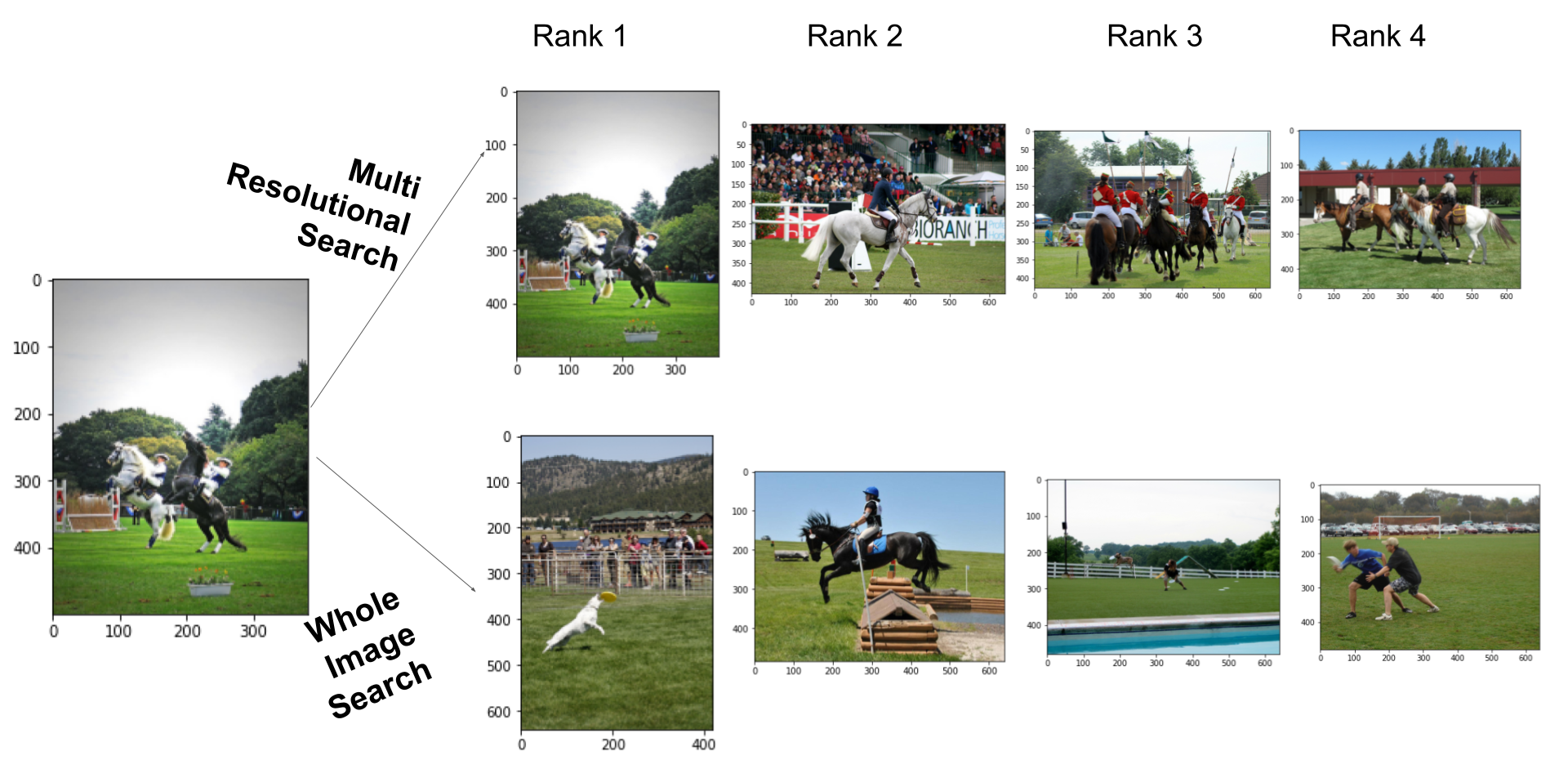} }}
    \caption{Cloud Removal and Multi-Resolution Image Search}
    \label{fig:data-preprocessor2}
\end{figure*}

\subsection{Data Downloader}
Our data source can be a vast stream of unlabelled images, but that data cannot be directly used to train machine learning models due to the lack of compute and storage. Frameworks also require datasets to adhere to a specified format. The Data Downloader helps source limited data from the data source and converts it into a format that can be directly used by the model training framework. Curator allows full flexibility to the user in defining the Data Source and the Data Downloader based on their domain.

\subsection{Data Preprocessor}
Data preprocessing is a fundamental data operation in ML that helps improve the model’s performance. Data present in certain domains like satellite imagery, medical imaging, and the like come with inherent discrepancies. Data Preprocessor consists of a set of statistical and geometric functions that were designed to solve some inherent deficiencies present in the satellite imagery data. These challenges are specific to the dataset. For instance, the NASA Worldview data had some esoteric deficiencies that we had to fix in order to make the data usable. 

\subsubsection{Cloud Removal}

Clouds are a major barrier in Remote Sensing datasets since they occlude the information of the space underneath. They make learning representations much harder for Machine Learning models. 
We were able to retrieve a cloudless version of an area by performing Image Subtraction over multiple images of the same area across several days. 
Contrarily, we were also able to retrieve cloud masks out of images individually, which can greatly help with cloud segmentation problems (see Figure 4(a)).

\subsubsection{Swath-Fillers}
Image tiles retrieved from the Worldview MODIS product come with small gaps at the equator, called Swaths. These occur due to the nature of the movement of the satellite over the earth.
Training models on images containing Swaths meant an ML model learns this as a feature across images and clusters them together. This affects performance greatly.
Through the Nearest pixel interpolation strategy, we were able to perform a Content-Aware fill on these swaths with relevant surrounding information \cite{chen2021reducing} (see Figure 3). 
This problem has also recently been overcome by sourcing our data from another product named VIIRS on GIBS.



\subsubsection{Multi-Resolution Image Search}

Images in Remote Sensing datasets can appear in different resolutions. 
There can be images with the class object appearing in different sizes, as well as the presence of multiple objects in an image.
Similarity search precision can be affected due to this. 
By tiling the image into a grid of patches, and obtaining the nearest neighbors for each of those tiles, we were able to aggregate the results by using a bucket voting strategy. This helped put the embedding distances into context and return similar matches to the entire image based on the voted scoring(See Figure 4(b)). 
    Although in practice, we found that Multi-Resolution search was a time consuming process that struggled at scale, so instead we built a model store that consists of models trained on multiple resolutions. We utilize the corresponding model based on the resolution of the image being used.

\subsubsection{Diverse Data Sampler}

Data Imbalance is a real problem in Machine Learning. For instance, satellite imagery datasets are inherently biased due to the natural imbalance between the different classes present in them. 71\% of the tiles present consist of water bodies, and our ML systems find it hard to learn information about poorly represented classes such as those images of natural phenomena, due to their sheer lack of occurrence in the data.

We apply a coreset strategy to the data to obtain a more representative sample of our data. This was absolutely necessary since we had the resources to only train on a subset of our entire pool of satellite imagery data. In simpler terms, we pick the farthest point for the current set of points, until the set equals the sample size. The resulting embedding space is an equidistant set of points that represent a diverse subset. This diverse subset is believed to contribute more information to a model during training compared to a randomly sampled subset. The standard algorithm is a deterministic operation for a given starting point, and the number of operations done is \textit{subset\_size * subset\_size * total\_num\_samples}

For a more scalable version, we also introduce a stratified version of this sampler,where instead of going through the entire embedding space, this technique first samples a random set of points, determines the farthest point from that sample, resamples a new set of points and repeats the process until a diverse sample is obtained. Resampling is done periodically to prevent the selection of farthest points within a sample of the embedding space. Num operations done is \textit{subset\_size * subset\_size * num\_random\_samples}

While working with large scale satellite datasets, like the one from NASA Worldview, we found that it was extremely time consuming to perform a forward pass over all ~10 million tiled images from the dataset. Instead we employed the Diverse Data Sampler to pick a highly representative sample of just 10\% of the data, thereby significantly reducing the time taken to perform a forward pass. Overall, along with the aforementioned optimizations, there is potential to reduce the runtime from initally taking 21,000 hours to just 13 minutes with no degradation in model quality. 

\end{document}